# Infrared target tracking based on proximal robust principal component analysis method

**Chao Ma [1], Guohua Gu [1, *], Xin Miao [2], Minjie Wan [1], Weixian Qian [1], Kan Ren [1], and Qian Chen [1]**

1. School of Electronic and Optical Engineering, Nanjing University of Science and Technology, Nanjing 210094, China; machaonjust@njust.edu.cn (C.M.); minjiewan1992@njust.edu.cn (M.W.); qianweixian_njust@yahoo.com (W.Q.); k.ren@njust.edu.cn (K.R.); chenq@njust.edu.cn (Q.C.)

2. The 28th Research Institute of China Electronics Technology Corporation，Nanjing 210007, China; yudi6134_cn@sina.com.cn (X.M.)

* Correspondence: gghnjust@mail.njust.edu.cn; Tel.: +86-258-431-0993

**Abstract:** Infrared target tracking plays an important role in both civil and military fields. The main challenges in designing a robust and high-precision tracker for infrared sequences include overlap, occlusion and appearance change. To this end, this paper proposes an infrared target tracker based on proximal robust principal component analysis method. Firstly, the observation matrix is decomposed into a sparse occlusion matrix and a low-rank target matrix, and the constraint optimization is carried out with an approaching proximal norm which is better than $l_1$–norm. To solve this convex optimization problem, Alternating Direction Method of Multipliers (ADMM) is employed to estimate the variables alternately. Finally, the framework of particle filter with model update strategy is exploited to locate the target. Through a series of experiments on real infrared target sequences, the effectiveness and robustness of our algorithm are proved.

**Keywords:** infrared target tracking; proximal robust principal component analysis; particle filter framework; template update

## 1. Introduction

Target tracking has been a very important part in many areas, such as traffic monitoring, detecting of approaching targets, video surveillance and activity control. Compared to the target tracking using visible camera, which has been deeply investigated in the past several decades, infrared target tracking is a more suitable and effective method to work in the lightless environment. Besides，infrared system is more robust in various environments, whether in dark or light condition. Thus, infrared target tracking has been applied more and more widely in military and civil applications [1]. On this basis, we hold the opinion that it is necessary to invest a high-precision infrared target tracking algorithm in different occasions.

Although the infrared system can work continuously, regardless of day or night, the image information obtained from infrared imaging devices is not as sufficient as that from visible devices due to the imaging principle and technical bottleneck [2]. For example, infrared image is lack of color and texture information. What is more, there are sensor noises and occlusion clutters in infrared images. Thus, the loss of image details and low signal-to-noise ratio limit the development of infrared target tracking technology. In addition, partial occlusion or total occlusion are also technical problems in the process of tracking. In brief, it is vital to establish a precise and credible model for infrared target tracking.

Based on the review of the existing target tracking algorithms, we divide them into two kinds: generative approaches and discriminant approaches. The generative model first extracts the feature of the target and learn the feature model. Through matching the whole picture, the target that the most similar area can be found. In recent years, target detection algorithms based on Bayesian model have been extensively studied and achieved satisfactory performance. The kernel-based tracker is widely investigated, including Mean-shift (MS) tracker [3], which is one of the mainstream target tracking methods. This method uses the steepest descent algorithm to solve iteratively in the direction of gradient descent, the core of which is to iterate over the solution until the optimal value is reached. This process is also to find the candidate which is the most similar to the template. However, MS algorithm cannot track target with changeable size adaptively. To solve this problem, Allen [4] proposed a Cam-shift tracker, which is an improved strategy of MS. It adds scale adaptive mechanism on the original MS to make it adaptively change the size of the tracking box according to the size of the target. Isard et al. [5] firstly investigated a condensation algorithm which applied particle filter framework to target tracking problem for the first time. Nevertheless, due to the lack of observation information at the current moment, the tracking performance turns to be poor in complex scenes. To work out the problems in the target tracking, such as poor tracking accuracy, large computation and poor real-time performance, researchers have proposed many improved particle filter-based tracking algorithms During the past decades. Guo et al. [6] used the particle filter algorithm to predict the initial target position, in which whether the real-time model of the target is changed or not is decided according to the similarity between the real target model and the predicted target model. In order to overcome the difficulty caused by the change of appearance, incremental visual tracking (IVT) [7] were proposed, which uses incremental algorithm based on principal component analysis.

Discriminant approaches consider both the background and the target, and extract the target model by comparing the differences between the two parts, so that the position of the current frame can be predicted. Zhang et al. [8] applied the theory of compressed sensing (CS) to the target tracking and proposed a fast compression tracking (FCT) [9] method. Babenko et al. [10] proposed an online learning tracker, which conducts online training for the classifier, and put positive samples and negative samples into two different parts. Furthermore, a weight multiple instance learning (WMIL) tracker was developed [11] by adding sample weights into the learning process. Zhang et al.[12] proposed a simple and effective algorithm for online feature selection (ODFS). The objective function is optimized iteratively along the steepest ascending gradient of the positive sample and the steepest descending gradient of the negative sample to maximize the output of the weak classifier. A new online visual tracking method based on probability continuous outlier model (PCOM) was presented by Wang et al. [13], and achieves very good performance in both accuracy and speed.

As deep learning emerges in more and more areas, researchers have started to employ it to target tracking. At present, most of the target tracking using deep learning belong to the discriminant approaches. Different from the significant role of deep learning in fields like object detection and recognition, the development of deep learning in the field of target tracking is not smooth, mainly due to the lack of training data. One of the core strengths of deep learning is the effective training of a large number of samples, whereas, target tracking only uses the first frame for sample training before starting. Wang et al. [14] proposed a deep learning tracker (DLT), which applies the methods of offline pre-training and online fine-tuning to neural network training, largely solving the problem of lack of samples. Zhang et al. [15] proposed a convolutional network (CNT) tracker. Compared with the traditional method based on deep learning, CNT does not require a large sum of data for pre-training.

Although the existing target tracking methods have made great progress, the interferences of target overlap, pose change; and environmental occlusion still need to be further studied and overcome [16]. Xue et al. [17] introduced sparse representation into the field of target tracking for the first time, and transformed the tracking into solving an $l_1$ minimization problem, which can adapt to the interferences such as pose change, severe occlusion and illumination change. Since then, generative models based on low rank and sparse representation have been proposed. Its basic principle is that the target matrix composed of multiple similar groups of target images possess low-rank property. Besides, the occlusion matrix representing partial occlusion is sparse since its entries only occupy a small number of the image. On this basis, this paper proposes a tracker based on proximal robust principal component analysis-based tracker. Firstly, the observation matrix is decomposed into a sparse occlusion matrix and a low-rank target matrix, and the constraint optimization is carried out with an approaching proximal norm. In order to solve this convex optimization problem, ADMM [18] algorithm is introduced to solve the variables. Finally, this paper uses the Bayesian state inference and model update mechanism under the particle filter framework for long-term tracking. Through a series of experiments on real infrared target sequences, the effectiveness and robustness of our algorithm are proved.

In conclusion, the main contributions of this paper can be summarized as follows:
(1) An appearance model based on sparse representation and low-rank is proposed to solve the tracking problem by transforming into a convex optimization problem;
(2) Proximal *p*-norms which declines to zero rapidly when the entry of the sparse matrix approaches zero is utilized to constrain occlusion matrix;
(3) A dynamic template update strategy is designed so that the appearance change can be well addressed.

The rest of the article is divided into five sections. Section 2 mainly introduces the related work of particle filter. In Section 3, we introduce the establishment of the model in detail. Section 4 demonstrates the reliability of the proposed algorithm through quantitative and qualitative experiments. The last section is a summary of the article.

## 2. Related Work

Particle filter is one of the most widely used methods based on random sampling in the field of target tracking. The tracking framework of particle filter can be regarded as a Bayesian inference problem, which contains two main steps: state update and state prediction. In the tracking process, the target state at time $t$ can be represented by the a 6-dimensional affine component $\mathbf{X}_t = (h_t, w_t, s_t, r_t, \theta_t, \lambda_t)^\mathrm{T}$ which is related to the initial time, where $h_t$ and $w_t$ represent the spatial position at time $t$, $s_t$, $r_t$, $\theta_t$ and $\lambda_t$ represent the scaling scale, rotation angle, aspect ratio and tilt angle. Assuming that $\mathbf{Y}_{1:t-1}=\{\mathbf{Y}_1,\mathbf{Y}_2,...,\mathbf{Y}_{t-1}\}$ is the target observation from the initial time to the current time $t$, we can use the iterative equation shown below to predict the prior probability distribution of the current target state $\mathbf{X}_t$:

$$p(\mathbf{X}_t|\mathbf{Y}_{1:t-1})=\int p(\mathbf{X}_t|\mathbf{Y}_{t-1})p(\mathbf{X}_{t-1}|\mathbf{Y}_{1:t-1})d\mathbf{X}_{t-1}, \tag{1}$$

where, $p(\mathbf{X}_t|\mathbf{Y}_{1:t-1})$ is the transfer probability of the target state, and obeys the Gaussian distribution:

$$p(\mathbf{X}_t|\mathbf{X}_{1:t-1})\sim N(\mathbf{X}_{t-1};\mathbf{\Sigma}^2), \tag{2}$$

where, $\mathbf{\Sigma}^2 = \mathrm{diag}(\sigma_w^2,\sigma_h^2,\sigma_s^2,\sigma_r^2,\sigma_\theta^2,\sigma_\lambda^2)$ is the diagonal covariance matrix, and its element is the variance of the affine component $\mathbf{X}_t = (h_t, w_t, s_t, r_t, \theta_t, \lambda_t)^\mathrm{T}$ corresponding to the element parameter.

Since the observed $\mathbf{Y}_t$ is known at time $t$, the posterior probability density can be given by the Bayes rule:

$$p(\mathbf{X}_t|\mathbf{Y}_{1:t})=\frac{p(\mathbf{Y}_t|\mathbf{X}_t)p(\mathbf{X}_t|\mathbf{Y}_{1:t})}{p(\mathbf{Y}_t|\mathbf{Y}_{1:t-1})}, \tag{3}$$

where, $p(X_t|Y_{1:t})$ is the likelihood probability to measure the similarity between the observed quantity $\mathbf{Y}_t$ and the actual target. Considering $p(X_t|Y_{1:t})$ is a constant, we can get

$$p(X_t|Y_{1:t}) \propto p(Y_t|X_t)p(X_t|Y_{1:t-1}), \tag{4}$$

At the current time t, the target state can be gotten by solving the maximum likelihood probability $p(X_t|Y_{1:t})$ as Eq. (5).

$$\mathbf{X}_t^* = \arg\max_{X_t} p(\mathbf{X}_t | \mathbf{Y}_t), \tag{5}$$

## 3. Model Establishment

In this section, we mainly introduce the infrared target tracking model based on the approach to proximal principal component analysis. The main symbol meanings used in the model are explained in Section 3.1; the appearance model is established in Section 3.2; and the calculation algorithm is given in Section 3.3. In view of the change of target shape and attitude, a template update strategy is proposed in Section 3.4. in Section 3.5, the algorithms in this section are summarized.

3.1 Definitions

Before the start of tracking, target region of the image is manually selected, and perturbations are added to the target region to form $i$ region of objects image. Their columns are vectorized, thus forming the template matrix $\mathbf{F}=\{\mathbf{f}_1,\mathbf{f}_2,...\mathbf{f}_i\}_{j\times i}(j \gg i)$, where $f_i$ represents the $i$-th column vector of the template matrix and $j = w \times h$. The candidate target $m_k \in \mathbf{R}^{j \times 1}$, which is normalized from $w \times h$, and the template matrix $\mathbf{F}$ together form the observation matrix $\mathbf{M} = \{\mathbf{F}, m_k\} = \{f_1, f_2, ... f_i, m_k\} \in R^{j \times (i+1)}$. $\mathbf{L}$ is the target matrix and has the same dimension as $\mathbf{M}$. In other words, each column of $\mathbf{L}$ represents an estimate of the corresponding target pixel in $\mathbf{M}$. In this paper, $\mathbf{S} = \{s_1, s_2, ..., s_i, s_{i+1}\}$ is used as the occlusion matrix, where each column $s_i$ represents the estimation of the occlusion pixel in $\mathbf{M}_i$.

3.2 Appearance Model

The observation matrix $\mathbf{M}$ is broken down into sparse occlusion matrix $\mathbf{S}$ and low-rank target matrix $\mathbf{L}$. Then the target appearance model of the algorithm in this paper can be expressed as

$$\mathbf{M} = \mathbf{L} + \mathbf{S}, \tag{6}$$

For the purpose of reconstructing target observation matrix $\mathbf{M}$, the function is used to calculate the minimum reconstruction error, as follows

$$\min_{\mathbf{L},\mathbf{S}} \frac{1}{2} \|\mathbf{M} - \mathbf{L} - \mathbf{S}\|_F^2, \tag{7}$$

For a general infrared image sequence, the objects of adjacent frames have little difference in shape and gray scale and strong linear correlation, so these vectorized similar image regions, which are vectorized, can constitute a low-rank matrix [19]. Therefore, we can get the following constraint applied to the target matrix $\mathbf{L}$:

$$rank(\mathbf{L}) \leq \omega, \tag{8}$$

where, $rank(\cdot)$ is the rank operator and $\omega$ is a small constant. Since the nuclear norm can replace the rank operator [20], the target apparent model can be expressed as

$$\min_{\mathbf{L},\mathbf{S}} \frac{1}{2} \|\mathbf{M} - \mathbf{L} - \mathbf{S}\|_F^2 + \lambda \|\mathbf{L}\|_*, \tag{9}$$

where, $\lambda$ is the weight factor. According to robust principal component analysis (RPCA) [21], we set $\lambda = (1/10)\sqrt{\max(j, i+1)}$ in this paper.

In the practical infrared target tracking process, the change of illumination, partial occlusion or total occlusion, may lead to unpredictable positioning error. According to the prior hypothesis information, the occlusion area only occupies a small part of the selected image block, so the occlusion matrix is sparse and the proximal *p*-norm can be incorporated into the model [22].

Next, the proximal *p*-norm of the matrix is defined as follows.

$$G_{\mu,p}(\mathbf{S}) = \sum_{m,n} g_{\mu,p}(s_{m,n}), \tag{10}$$

where, $\mu$ is the parameter and $g_{\mu,p}$ is implicitly determined by

$$|s|^2/2 + \mu g_{\mu,p}(s) = (|\cdot|^2/2 - \mu h_{\mu,p})^*(s), \qquad (11)$$

where, the operator $*$ is the Legendre-Fenchel transformation to the function [23], and the $h_{\mu,p}$ is defined as

$$h_{\mu,p}(t) = \begin{cases} |t|^2/2\mu & \text{if } |t| \leq \mu^{\frac{1}{2-p}} \\ |t|^p/p - \delta & \text{if } |t| \geq \mu^{\frac{1}{2-p}}, p \neq 0, \\ \ln|t| - \dfrac{\ln\mu}{2} + \dfrac{1}{2} & \text{if } |t| \geq \mu^{\frac{1}{2-p}}, p = 0 \end{cases} \qquad (12)$$

where, $\delta = (1/p - 1/2)\mu^{p/(2-p)}$. When $p = 0$, $h_{\mu,0}(t)$ can be defined as the limit as $p$ approaches 0.

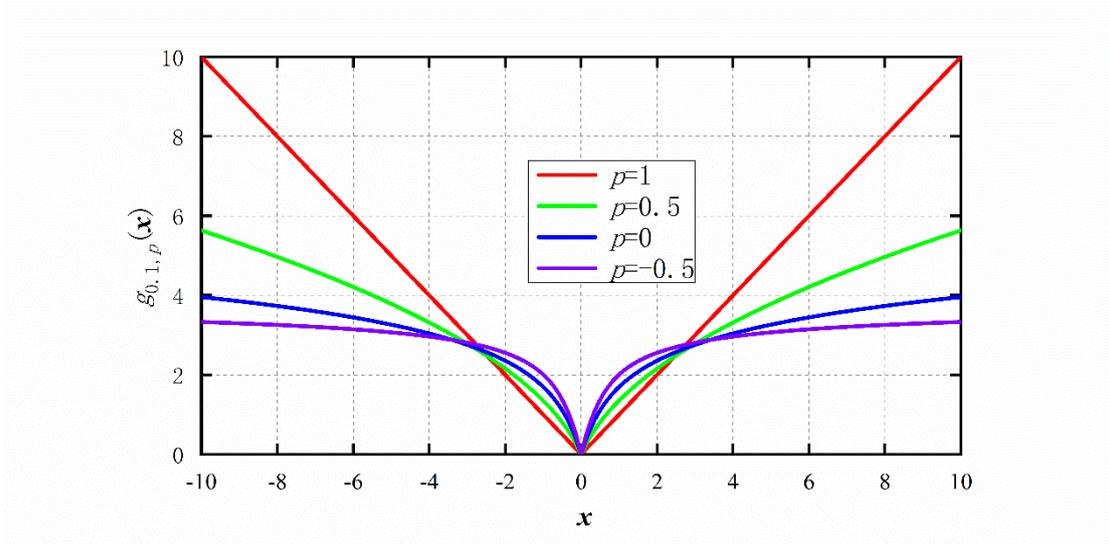

**Figure 1.** Curves of $g_{\mu,p}$ with $\mu = 0.1$ and different $p$.

As shown in the figure 1, the function $g_{\mu,p}$ and $l_p$-norm have similar characteristics. When $p < 1$, $g_{\mu,p}$ rapidly drops to 0 at $x = 0$, showing that the curve has a sharp edge at $x = 0$. Then, the matrix element is 0, the penalty function value is 0. What's more, if a matrix element deviates from $x = 0$, the value of the penalty function increases rapidly. At last, the value of the function is nearly unchanged, when the value of the matrix element is far from 0. Therefore, the proximal $p$-norm can well constrain the sparsity of the matrix. Finally, the target apparent model of the algorithm in this paper can be expressed as

$$\mathcal{L}(\mathbf{L},\mathbf{S},\mathbf{I};\mu) = G_{\mu,p}[\sigma(\mathbf{L})] + \lambda G_{\mu\lambda,p}(\mathbf{S}) + \frac{1}{2\mu}\|\mathbf{M}-\mathbf{L}-\mathbf{S}+\mathbf{I}\|_F^2, \qquad (13)$$

where, the kernel norm of matrix $\mathbf{L}$ is changed to proximal $p$-norm of its singular value matrix $\sigma(\mathbf{L})$. $\mathbf{I}$ is the Lagrange multiplier, $\mu$ is the scale factor, and $G_{\mu,p}$ is changed to $G_{\mu\lambda,p}$ for the convenience of subsequent solutions.

For each frame of image, N corresponding candidate targets can be obtained from modeling the state transition probability $p(\mathbf{X}_t | \mathbf{X}_{t-1})$ with gaussian distribution. The reconstruction error matrix $\varepsilon_t$ of each candidate target can be calculated by

$$\varepsilon_t = \mathbf{M}_{1:i+1} - \mathbf{L}_{1:i+1} - \mathbf{S}_{1:i+1}, \tag{14}$$

where, $\mathbf{M}_{1:i+1}$, $\mathbf{L}_{1:i+1}$, $\mathbf{S}_{1:i+1}$ respectively represent the observation matrix, target matrix and occlusion of the current candidate target $m_k$.

The likelihood probability $p(\mathbf{Y}_t | \mathbf{X}_t)$ can be calculated according to the reconstruction error:

$$p(\mathbf{Y}_t | \mathbf{X}_t) = \frac{1}{\sqrt{2\pi}} \prod_x \prod_y \exp(-\frac{\varepsilon_t(x,y)^2}{2\sigma_\varepsilon^2}), \tag{15}$$

Eq. (15) shows that $p(\mathbf{Y}_t | \mathbf{X}_t) \propto N(\varepsilon_t(x,y); 0, \sigma_\varepsilon^2)$ follows the gaussian distribution with the mean value of zero and the variance of $\sigma_\varepsilon^2$.

3.3 Solution

Eq. (13) is a multivariable optimization problem, which can be calculated by ADMM algorithm [24]. The ADMM algorithm solves one of the variables alternately and constantly updates the Lagrange multiplier, and the convergence of the Lagrange algorithm can be referred to [18]. The specific update step of ADMM algorithm is as follows

$$\begin{cases} \mathbf{S}_{t+1} = \arg \min_{\mathbf{S}} \mathcal{L}(\mathbf{L}_t, \mathbf{S}, \mathbf{I}_t; \mu_t) \\ \mathbf{L}_{t+1} = \arg \min_{\mathbf{L}} \mathcal{L}(\mathbf{L}, \mathbf{S}_{t+1}, \mathbf{I}_t; \mu_t) \\ \mathbf{I}_{t+1} = \mathbf{I}_t + \frac{1}{\mu_t}(\mathbf{M} - \mathbf{L}_{t+1} - \mathbf{S}_{t+1}) \\ \mu_{t+1} = \rho \mu_t, \ 0 < \rho < 1 \end{cases}, \tag{16}$$

The above equation can be decomposed into several univariate programming problems.

1) Estimation of the occlusion matrix $\mathbf{S}$. Eq. (13) is simplified to

$$\min_{\mathbf{S}} G_{\lambda\mu_t,p}(\mathbf{S}) + \frac{1}{2\lambda\mu_t} \| \mathbf{M} - \mathbf{L}_t - \mathbf{S} + \mathbf{I}_t \|_F^2, \tag{17}$$

The analytic solution of Eq. (17) is given through the $p$ contraction operator [18]:

$$\mathbf{S}_{t+1} = \mathcal{S}_p(\mathbf{M} - \mathbf{L}_t + \mathbf{I}_t; \lambda\mu_t), \tag{18}$$

2) Estimation of the target matrix $\mathbf{L}$. Similarly, Eq. (13) is simplified to:

$$\min_{\mathbf{L}} G_{\mu_t, p}(\sigma(\mathbf{L})) + \frac{1}{2\lambda\mu_t} \|\mathbf{M} - \mathbf{L} - \mathbf{S}_{t+1} + \mathbf{I}_t\|_F^2, \quad (19)$$

The analytical solution of Eq. (19) is given by SVD decomposition method:

$$\mathbf{L}_{t+1} = \mathbf{U} \mathcal{S}_p(\boldsymbol{\Sigma}, \mu_t) \mathbf{V}^T, \quad (20)$$

where, $\boldsymbol{\Sigma}$ is the singular value matrix of $\mathbf{M} - \mathbf{S}_{t+1} + \mathbf{I}_t$, namely $\mathbf{M} - \mathbf{S}_{t+1} + \mathbf{I}_t = \mathbf{U}\boldsymbol{\Sigma}\mathbf{V}^T$ [19].

3) Update of Lagrange multiplier $\mathbf{I}$ and scale factor $\mu$. According to Eq. (13), the decay factor $\rho$ is multiplied in each iteration, and $\mu$ is geometrically attenuated. When the convergence condition $\|\mathbf{M} - \mathbf{L} - \mathbf{S}_t\|_F / \|\mathbf{M}\| < 1e-5$ is satisfied, the iteration terminates and the detection result is obtained. The convergence and stability of the algorithm can be referred to [19] and [23].

3.4 Template Update Strategy

In the tracking process, both the tracking environment (such as occlusion, illumination, etc.) and the appearance of the target (such as scale, shape, etc.) may change, resulting in some templates being not applicable. Therefore, in case of significant deformation or occlusion mutation of the target, the elements in the template matrix T must be updated in time. However, the templates should not be updated too frequently. Otherwise the cumulative tracking error will be exacerbated. On the other hand, it also cannot be too slow in order to avoid the drifting from ground truth. It is worth mentioning that the template matrix cannot be updated when the target is occluded [25].

We use the reconstruction error $\tilde{\varepsilon}_k$ of the current frame tracking results $\tilde{y}$ to calculate the importance weight $w_k$, and give the weights to each element of the template matrix $\mathbf{F}_{1:i}$. Specifically, the iteration strategy as $w_k = w_k \cdot \exp(-\|\tilde{\varepsilon}_k(:,k)\|_F)$ is adopted, where $w_k$ is initialized to $w_k = 1/i$ in the first frame. In addition, the sum of the $(i+1)$-th column vector $\text{sum}(\mathbf{S}_{i+1})$ in the current occlusion matrix $\mathbf{S}$ is applied to measure the occlusion. If $\text{sum}(\mathbf{S}_{i+1})$ is above the threshold $\xi^* * j$, ($\xi^* \in [0,1]$), we argue that there are obvious occlusion in the current frame, thus the template matrix T cannot be updated. In summary, the template update steps are summarized in Algorithm 1.

In this paper, cosine similarity, the included angle between vectors, is applied to express the similarity between two vectors. Obviously, the greater the cosine similarity, the greater the difference between the current frame tracking result and the elements in the template matrix. According to the template update strategy, we replace the template of the minimum weight in with the current tracking result when the difference between the tracking result and the template matrix template is larger than a certain value. Of course, there is also the premise that occlusion is not too serious. In addition, the weight of the newly added template is established to the median value within all the templates, to prevent it from occupying the absolute dominance and affecting the tracking ability.

**Algorithm 1.** Template update strategy

**Input**: the template matrix T, the pre-defined thresholds $\psi^*$ and $\xi^*$.

1. **Initialize**: $w_k = \frac{1}{i}(k=1,2,...,i)$, where $i$ is the number of templates.
2. **For** t = 2, 3, ...
3. Calculate the current tracking result $y_k$, the corresponding target matrix $\mathbf{L}$, the occlusion matrix $\mathbf{S}$, and then calculated the reconstruction error $\tilde{\varepsilon}_k$ based on Eq.(14).
4. Update the template weight as $w_k = w_k \cdot \exp(-||\tilde{\varepsilon}_i(:,k)||_F)$.
5. Calculate the cosine similarity as $\psi_k = \arccos\langle \tilde{m}, f_k \rangle$.
6. Select the minimum value $\psi_{\min}$ in cosine similarity.
7. **If** $\psi_{\min} > \psi^*$ and $\text{sum}(s_{n+1}) > (\xi^* \cdot m)$
8. Replace the least weighted template in F with $\tilde{m}$, i.e. $f_c = \tilde{m}$, where $c = \arg\min_{1 \leq k \leq i} w_k$.
9. The new template is assigned a new weight $w_k = \text{median}(w)$, where $\text{median}(\cdot)$ represents the median value.
10. **End If**
11. Normalized template weight $w_k = w_k / \text{sum}(w)$.
12. Adjust $w$ to insure that max（$w$）=0.3.
13. **Output**: updated template matrix F and corresponding weight $w_k$.
14. **End for**

3.5 Summary of the Proposed Tracker

In this section, we summarize the proposed infrared target tracking algorithm based on proximal robust principal component analysis method.

**Algorithm 2.** The target tracking with complete steps

1. **Input**: the target box selected manually at the first frame.
2. **Initialize**: target state vector $\mathbf{X}_1$, template matrix $\mathbf{F}=\{f_1,f_2,...f_i\}\in R^{j\times i}$.
3. **For** $t=2$ to the last frame
4.     **For** $k=1$ to the last particle do
5.         Collect candidate sample $m_k \in R^{j\times 1}$, forming observation matrix $\mathbf{M}=\{f_1,f_2,...f_i,m_k\}\in R^{j\times (i+1)}$.
6.         **Initialize**: $\lambda = (1/10)\sqrt{\max(j,i+1)}$.
7.         **Loop**
8.         L-step: $\mathbf{L}_{t+1} = \arg\min_{\mathbf{L}} \mathcal{L}(\mathbf{L},\mathbf{S}_{t+1},\mathbf{I}_t;\mu_t)$.
9.         S-step: $\mathbf{S}_{t+1} = \arg\min_{\mathbf{S}} \mathcal{L}(\mathbf{L}_t,\mathbf{S},\mathbf{I}_t;\mu_t)$.
10.        Y-step: $\mathbf{I}_{t+1} = \mathbf{I}_t + \frac{1}{\mu_t}(\mathbf{M}-\mathbf{L}_{t+1}-\mathbf{S}_{t+1})$.
11.        **Until converge**
12.     **End For**
12.     Utilize the likelihood probability $P(\mathbf{Y}_t|\mathbf{X}_t)$ to resample.
13.     **Output**: $\mathbf{X}_t^* = \arg\max_{\mathbf{X}_t} p(\mathbf{X}_t|\mathbf{Y}_t)$.
14. **End for**

## 4. Experimental Results

In this section, 6 groups of typical infrared sequences are utilized to test the the proposed infrared target tracking algorithm on the MATLAB 2018b software platform. Then, the algorithm is compared quantitatively and qualitatively with other typical tracking algorithms.

4.1 Datasets

The interference in 6 test sequence include pose variation, scale change, occlusion and so on. The sequences can be downloaded from [26,27]. Details of each sequence are shown in Table 1, and the target of the first frame is manually selected (marked with red rectangle), as shown in Figure 2.

Table 1. Information of test sequence

| Database | Target | Background | Image Size | Challenge | Frame Number |
|---|---|---|---|---|---|
| Seq.1 | man | trees+ground | 640×480 | occlusion | 665 |
| Seq.2 | rhino | trees+ground | 320×256 | occlusion+pose change | 619 |
| Seq.3 | man | trees | 640×480 | overlap | 301 |
| Seq.4 | horse | ground | 608×480 | pose change | 348 |
| Seq.5 | man | ground | 640×480 | scale change+ overlap | 322 |
| Seq.6 | man | trees+ground | 320×240 | overlap | 449 |

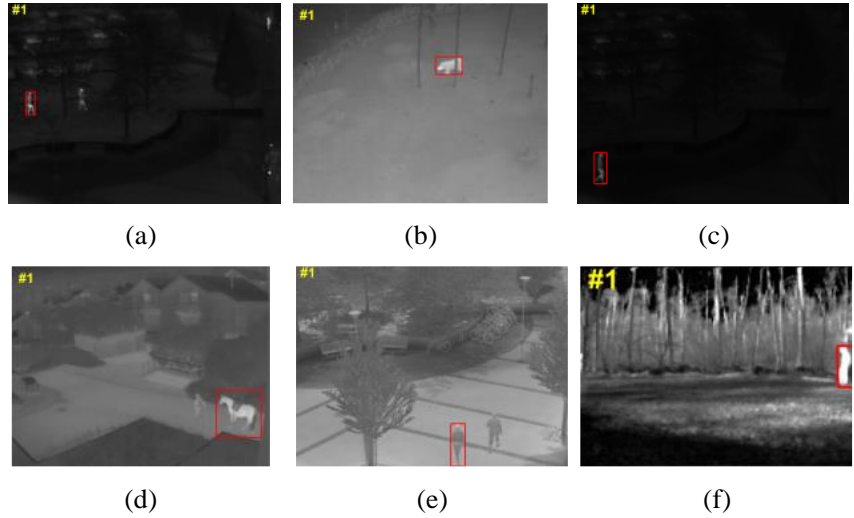

(a)　　　　　　　　　(b)　　　　　　　　　(c)

(d)　　　　　　　　　(e)　　　　　　　　　(f)

**Figure 2.** First frame of each sequence (the target is marked in red): (a) man; (b) rhino; (c) man; (d) horse; (e) man; (f) man

4.2 Parameter setting

Table 2 lists all the parameter values involved in this algorithm, as well as their corresponding meanings and default values.

Table 2. List of parameter values

| Parameter | Meaning | Default Value |
|---|---|---|
| n | template number | 10 |
| p | particle number | 500 |
| $\xi^*$ | Occlusion discrimination threshold | 0.1 |
| $\psi^*$ | cosine similarity threshold | 30 |

4.3 Results

In our proposed tracking algorithm, the number of templates and the number of particles are set to 10 and 500. Five results of each sequence images are shown in Figure 3.

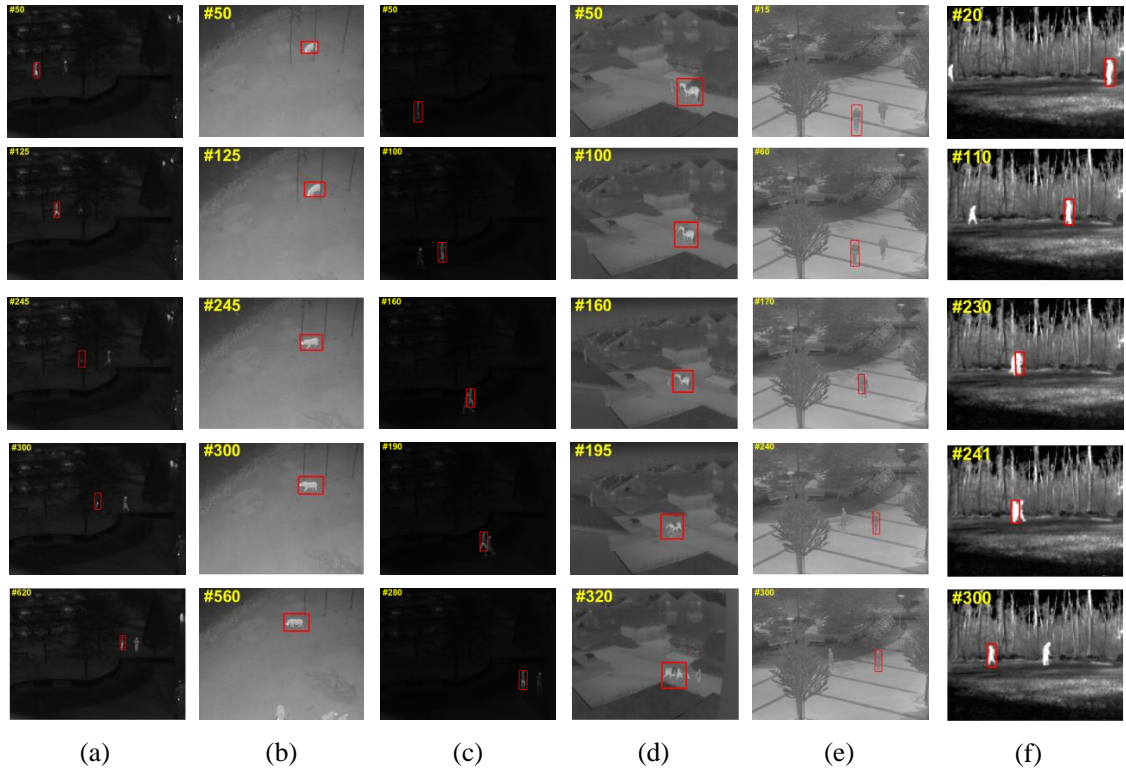

|  (a) | (b) | (c) | (d) | (e) | (f) |

Figure 3. The tracking results of our algorithm: (a) man; (b) rhino; (c) man; (d) horse; (e) man; (f) man

In Figure 3 (a), the target does not change significantly in appearance, nor does the template. The difficulty of this sequence is that around 245-th frame, the target is partially or completely obscured by trees. This problem is easy to solve because of the occlusion matrix in our algorithm. The size and posture of the rhino in Seq.2 are constantly changing, and there is also partial occlusion. Owing to the occlusion matrix and template update strategy, it has good tracking result. In Seq.3, the difficulty lies in the encounter with another person when the man is moving, which causes the remarkable overlap. Considering the low-rank property of the target matrix and the sparsity of the occlusion matrix, this kind of tracking performance is well. In addition, Seq.4 and Seq.5 are similar to Seq.2, with scale change and pose change. Seq.6 is similar to Seq.3 in which there is target overlap.

4.4 Qualitative Comparisons

This section presents the algorithm compared with 9 kinds of conventional algorithms, these algorithms including FCT [9], CNT [16], DLT [12], IVT [7], Meanshift (MS) [3], ODFS [12], $L_1$APG [17], PCOM [13], and WMIL [11]. The parameter values of all comparison algorithms are selected as their default parameters, and the particle numbers of DLT, CNT, IVT and $L_1$APG based on particle filter framework are uniformly set to 500. Figure 3 shows part of the tracking results.

FCT, ODFS and WMIL use haar-like features to represent the local information of the target. As shown in figure 3 (c, f). the algorithms do not track well when there is target aliasing in Seq.3 and Seq.6.

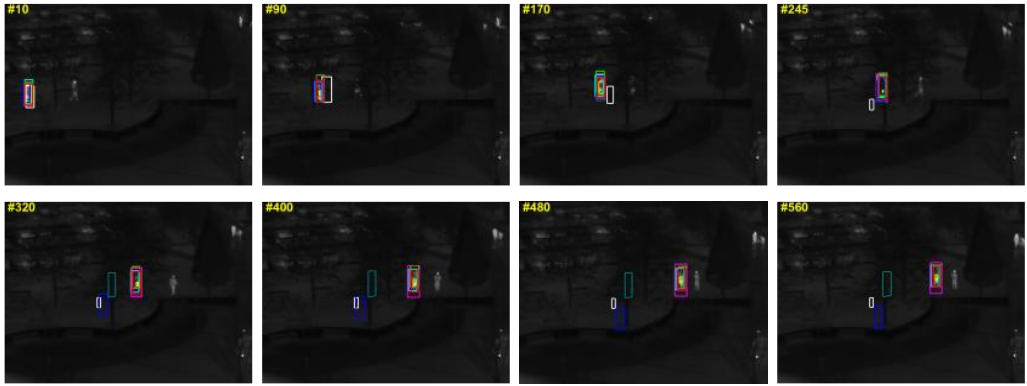

(a)

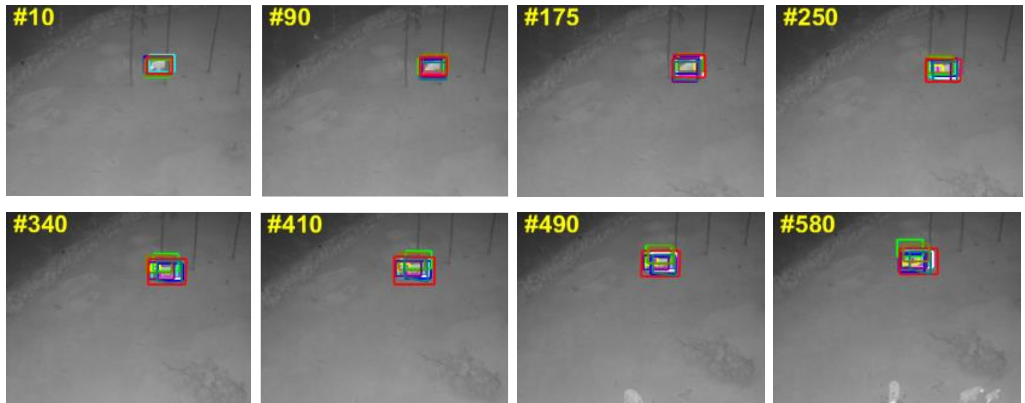

(b)

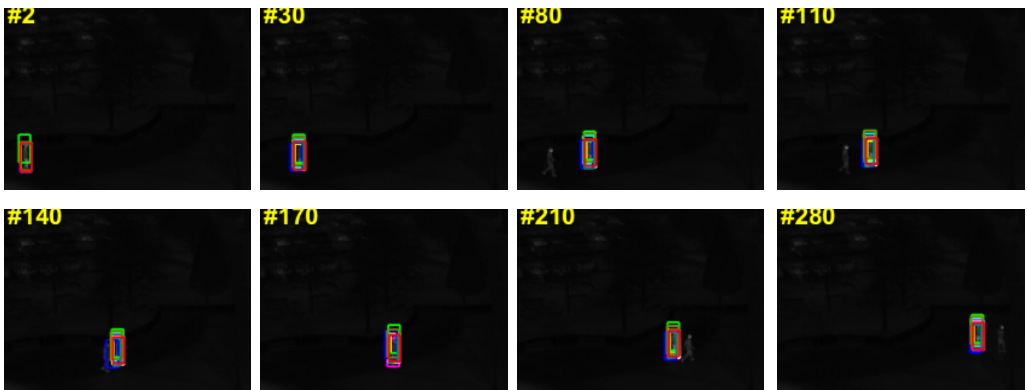

(c)

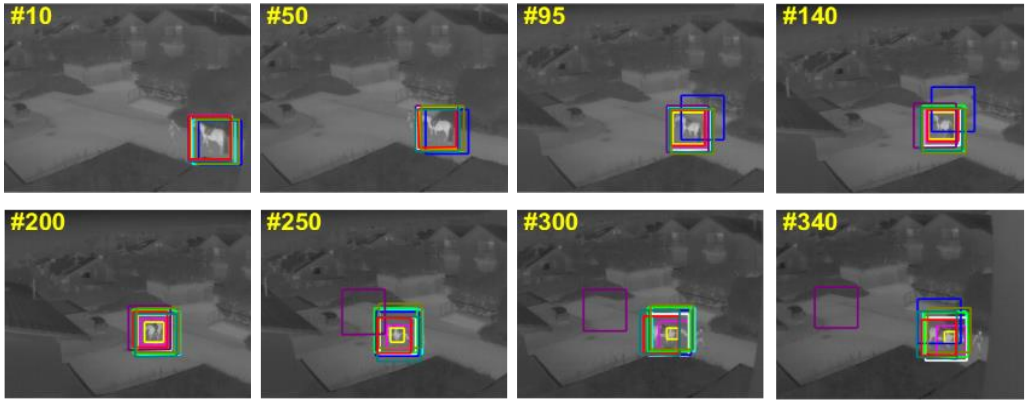

(d)

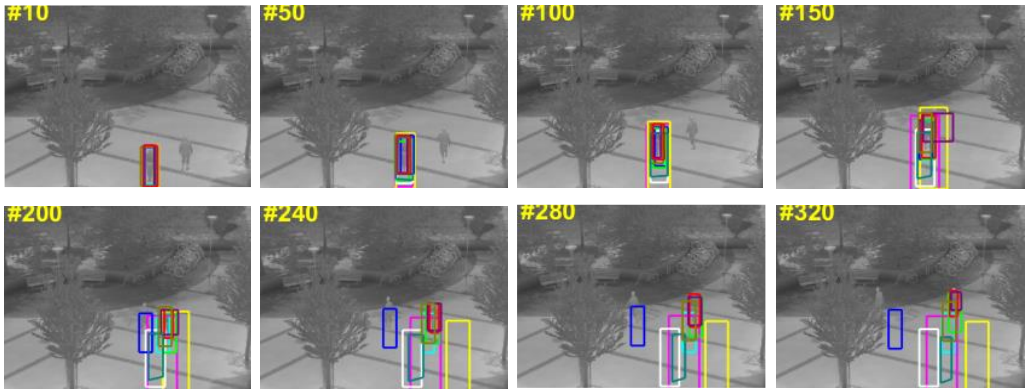

(e)

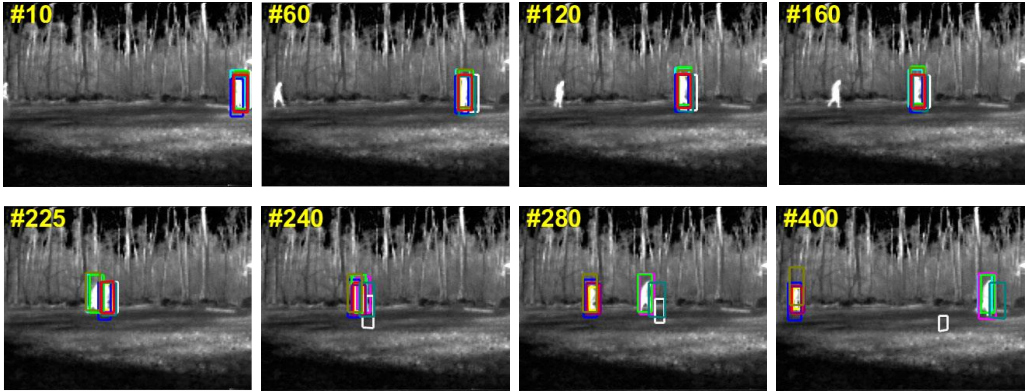

(f)

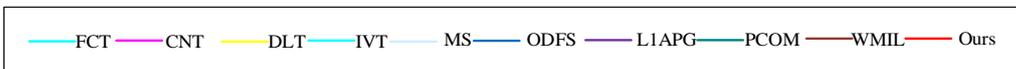

Figure 4. Tracking results of the 10 tracking algorithms over datasets: (a) man; (b) rhino; (c) man; (d) horse; (e) man; (f) man

MS and FCT can only achieve good tracking effect when the change of target's appearance is not drastic and there is not too much shielding. Because these two algorithms only take pixel gray level as the feature of the object appearance modeling, it is obvious that they cannot deal with relatively complex tracking scenes.

PCOM can utilize PCA subspace to estimate the change of target's appearance. As the figure shows that the algorithm achieves satisfactory tracking effect in Seq.3 and Seq4. If the adjacent occlusion pixel is misjudged as the main component while the target pixel is identified as a singular value, the algorithm will drift seriously.

The $L_1$APG uses the fragment template to represent the interferences such as occlusion and noise, so it performs stably in most test sequences (such as Seq.1, Seq.2, Seq.3 and Seq.6). Due to the random distribution of non-zero elements in the fragment template, the optimization equation based on $l_1$–norm may converge to the wrong value, resulting in tracking failure in some scenarios, such as Seq.4 and Seq.5.

DLT and CNT are convolutional network-based tracking algorithms which need to train the dataset before tracking. Compared with other comparison algorithms, the tracking results of these two algorithms are more accurate in general. However, DLT drifts on Seq. 3 and Seq.5, indicating that it is not stable for the overlap and appearance variation. CNT performs well except for Seq.5 and Seq.6. when the target size changes greatly, and the target aliasing is unstable.

On the contrary, the performance of the proposed tracker is stable, and it can track the target accurately in all the selected sequences. Experimental results shown above demonstrate that it is robust to occlusion, deformation, scale change and target overlap when compared to other traditional methods.

4.5 Quantitative Comparisons

In this section, we take two metrics to quantify the tracking results normalized. There are normalized center position error and average overlap score (AOS) [28,29].

The normalized center position error $\varepsilon_0$ is defined as:

$$\varepsilon_0 = \frac{\sqrt{(x_t - x_0)^2 + (y_t - y_0)^2}}{L_0} \qquad (21)$$

where, $(x_t, y_t)$ is the central coordinate of the tracking result, $(x_0, y_0)$ is the central coordinate of the ground truth, and $L_0$ is the diagonal length of the truth rectangular box.

The average normalized center position error $\overline{\varepsilon_0}$ is listed in Table 3. At the same time, the precision plot introduced in Figure 5 intuitively describes the tracking precision, which is defined as the percentage of image frames in which the center position error is less than a specific value.

Table 3. The average normalized center position error $\overline{\varepsilon_0}$

|  | FCT | CNT | DLT | IVT | MS | ODFS | WMIL | L$_1$APG | PCOM | Ours |
|---|---|---|---|---|---|---|---|---|---|---|
| Seq.1 | 0.0749 | 0.0324 | 0.1436 | 1.7798 | 1.6545 | 0.1415 | 0.1531 | 0.0638 | 0.1069 | 0.0564 |
| Seq.2 | 0.0757 | 0.2371 | 0.1338 | 0.1785 | 0.1506 | 0.2605 | 0.2434 | 0.0637 | 0.1792 | 0.1099 |
| Seq.3 | 0.1314 | 0.0320 | 0.0655 | 0.0207 | 0.1252 | 0.2480 | 0.1770 | 0.0210 | 0.0629 | 0.0236 |
| Seq.4 | 0.10583 | 0.0599 | 0.0558 | 0.0341 | 0.2152 | 0.0885 | 0.1427 | 0.5369 | 0.0846 | 0.0772 |
| Seq.5 | 0.3295 | 0.8714 | 0.7897 | 0.8657 | 0.6845 | 0.1957 | 0.1806 | 0.1381 | 0.8103 | 0.0432 |
| Seq.6 | 2.0589 | 2.0386 | 0.0579 | 1.9819 | 0.1045 | 2.0852 | 0.2206 | 0.0401 | 2.1962 | 0.0286 |
| Ave. | 0.4627 | 0.5452 | 0.2077 | 0.8101 | 0.4890 | 0.5032 | 0.1862 | 0.1439 | 0.5733 | 0.0564 |

As shown in Table 3 and Figure 5, the algorithm in this paper has the smallest overall center error in the six test sequences and is more ahead of other algorithms. The IVT and MS tracking results deviate from the truth value due to the severe occlusion in Seq.1. In Seq.2, all algorithms performed well and no targets were lost. In Seq.3, some tracking algorithms based on haar-like characteristics show different degrees of deviation after target overlap. The reason why L$_1$APG did not perform well in Seq.4 is that the posture of this group of sequences changed too much. In Seq.5, except the algorithm in this paper, all other algorithms lose targets to varying degrees, which is causing by scale changes and target overlap in this group of sequences at the same time, but this is what our algorithm is good at. In the last set of sequences, FCT, CNT, IVT, ODFS and PCOM algorithms all cause target loss after target overlap [30].

Unlike $\overline{\varepsilon_0}$, AOS further introduces the consideration of target scale when evaluating algorithm tracking progress. Assuming that the target boundary box obtained from the algorithm is $R_p$, and the boundary box of the target truth box is $R_p$, then AOS is defined as

$$\text{AOS} = \frac{|R_p \cap R_q|}{|R_p \cup R_q|} \tag{22}$$

Table 4 lists the average values of AOS on each sequence. At the same time, figure 6 shows the success plot of each algorithm on the test sequence. The abscissa is the AOS value between [0,1], and the ordinate is the percentage of image frames corresponding to AOS. As AOS increases from 0 to 1, the success rate decreases to 0.

Table 4. The average values of AOS

|       | FCT    | CNT    | DLT    | IVT    | MS     | ODFS   | WMIL   | $L_1$APG | PCOM   | Ours   |
|-------|--------|--------|--------|--------|--------|--------|--------|----------|--------|--------|
| Seq.1 | 0.7026 | 0.5676 | 0.2120 | 0.0721 | 0.2427 | 0.6173 | 0.6008 | 0.7004   | 0.3228 | 0.7217 |
| Seq.2 | 0.7267 | 0.4792 | 0.4071 | 0.5772 | 0.6079 | 0.3799 | 0.3812 | 0.7635   | 0.5518 | 0.5311 |
| Seq.3 | 0.5942 | 0.8499 | 0.5181 | 0.8965 | 0.5184 | 0.5139 | 0.5445 | 0.8843   | 0.6902 | 0.8819 |
| Seq.4 | 0.5572 | 0.7738 | 0.5519 | 0.7076 | 0.4182 | 0.5570 | 0.5204 | 0.3817   | 0.5834 | 0.7338 |
| Seq.5 | 0.3980 | 0.2197 | 0.1944 | 0.2179 | 0.3039 | 0.5259 | 0.4587 | 0.5145   | 0.2281 | 0.7664 |
| Seq.6 | 0.3245 | 0.4275 | 0.6437 | 0.2281 | 0.5814 | 0.3434 | 0.5299 | 0.7367   | 0.322. | 0.8451 |
| Ave.  | 0.5505 | 0.5530 | 0.4212 | 0.4498 | 0.4454 | 0.4896 | 0.5059 | 0.6635   | 0.4497 | 0.7467 |

The algorithm in this paper has achieved an excellent success rate in each test sequence and is obviously superior to other algorithms on the whole. In Seq.1, the CNT and DLT performed poorly in terms of success rate compared to excellent performance in accuracy, because such algorithms track local information about the target rather than the whole. ODFS and WMIL have a low success rate on Seq.2 due to the change of target pose and partial occlusion. All algorithms can track the target on Seq.3, but the success rate of DLT and MS is low. There is target pose change in Seq.4, and the poor performance of MS and $L_1$APG lies in that these two algorithms cannot deal with such problems as pose change. In Seq.5, scale change and target overlap occur simultaneously, and only the algorithm in this paper achieves excellent results. Other algorithms offset the target to different degrees. In the last sequence, FCT, CNT, IVT, and ODFS track the wrong targets with poor success rates due to target overlap.

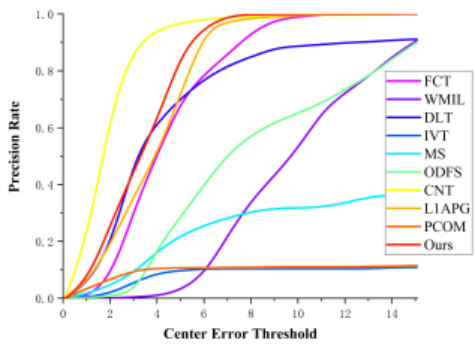

(a)

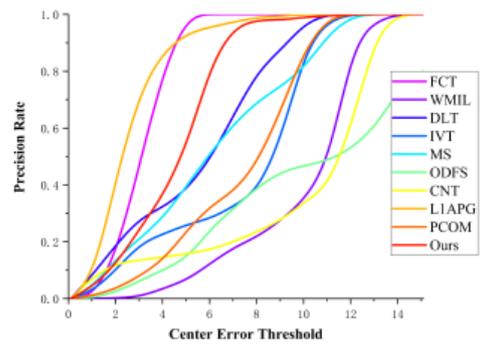

(b)

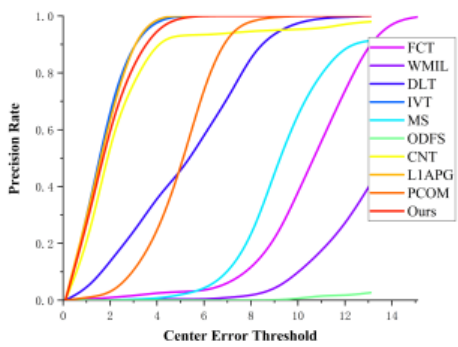

(c)

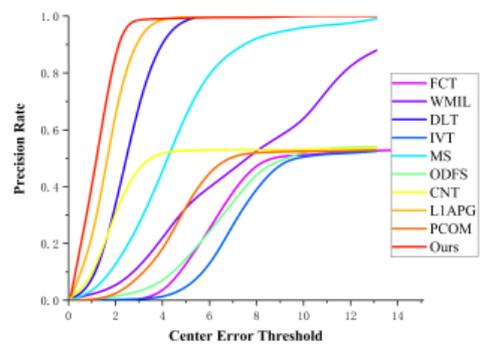

(d)

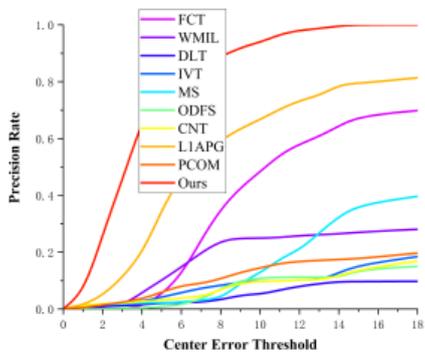

(e)

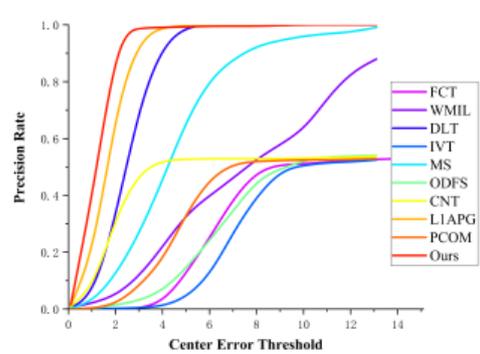

(f)

**Figure 5.** Precision plots: (a) man; (b) rhino; (c) man; (d) horse; (e) man; (f) man

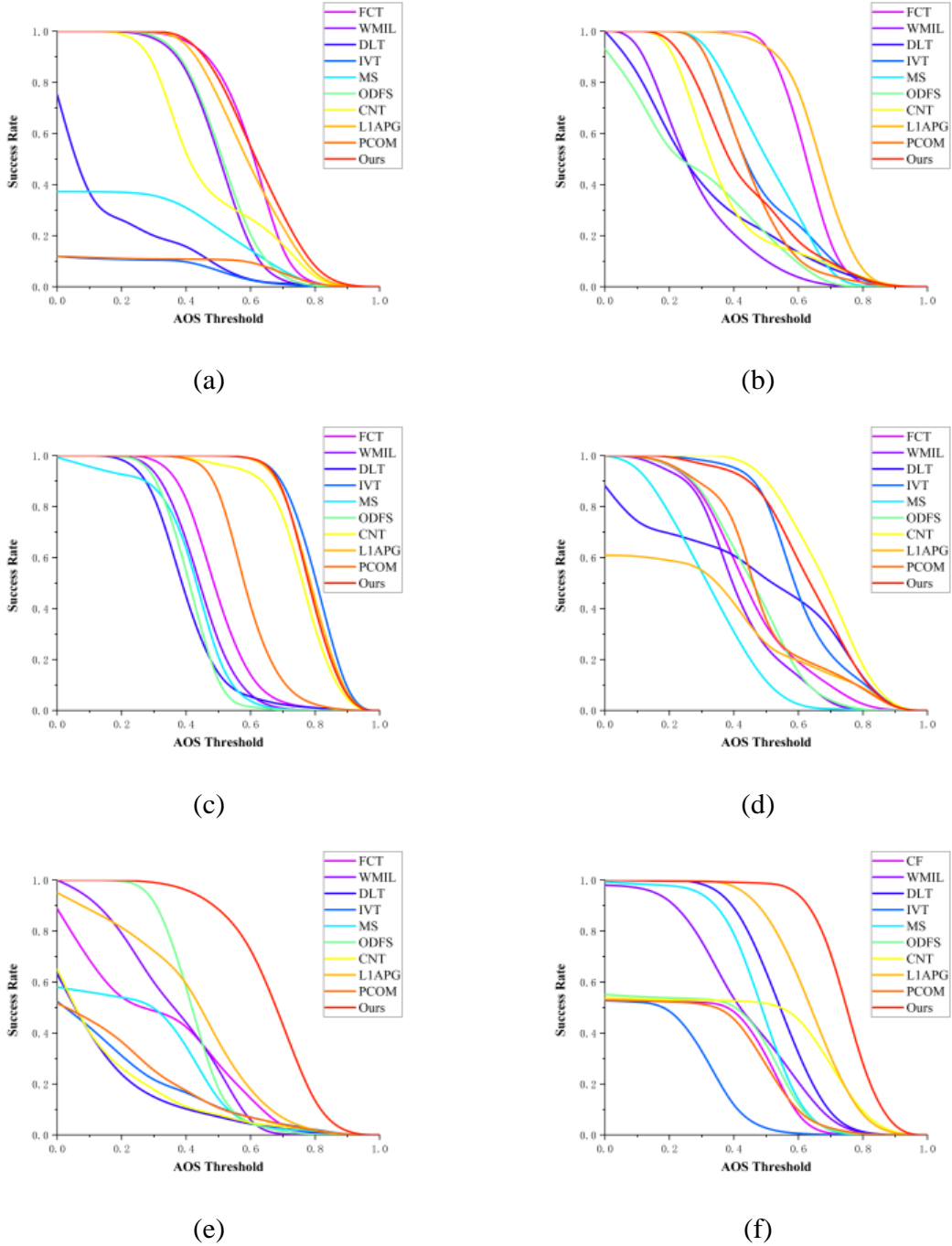

**Figure 6.** Success plots: (a) man; (b) rhino; (c) man; (d) horse; (e) man; (f) man

## 5. Conclusions

A target tracker based on approximate robust principal component analysis is proposed in this paper. Firstly, the observation matrix is decomposed into a sparse occlusion matrix and a low-rank target matrix, and the constraint optimization is carried out with an approximate norm that is closer than the approximation. In order to solve this convex optimization problem, ADMM algorithm is used to solve one variable alternately. For long-term tracking, this algorithm uses the Bayesian state inference and model update mechanism under the particle filter framework. Through a series of experiments on real infrared target sequences, the effectiveness and robustness of the algorithm are verified, and it is better than other advanced algorithms.

Although the algorithm has achieved excellent results in testing infrared sequences, there are still some interferences, such as non-rigid motion and long-term occlusion, which deserve further study. To cope with these disturbances, we plan to take other measures, such as improving appearance model and introducing motion estimation. At the same time, we hope to port MATLAB code to GPU or FPGA, to achieve real-time tracking.